\documentclass[manuscript]{acmart}
\usepackage{times,latexsym}
\usepackage{url}
\usepackage[T1]{fontenc}
\usepackage{amsmath}
\usepackage{xcolor}
\usepackage{tcolorbox}
\usepackage{soul}
\usepackage{comment}
\usepackage{graphicx}
\usepackage{multirow} 
\usepackage{ulem}
\usepackage{xspace,mfirstuc,tabulary}

\AtBeginDocument{%
  }

\begin{document}


\title{Detecting Voice Phishing with Precision: Fine-Tuning Small Language Models}

\author{Ju Yong Sim}
%\authornotemark[]
%\email{}
\author{Seong Hwan Kim}
\authornote{Corresponding author}
\email{seonghwan.kim@ut.ac.kr}
\orcid{1234-5678-9012}
\affiliation{%
  \institution{Korea National University of Transportation}
  \city{Uiwang}
  \state{Gyeonggi-do}
  \country{Republic of Korea}
}

\renewcommand{\shortauthors}{J. Y. Sim et al.}

\begin{abstract}
We develop a voice phishing (VP) detector by fine-tuning Llama3, a representative open-source, small language model (LM). In the prompt, we provide carefully-designed VP evaluation criteria and apply the Chain-of-Thought (CoT) technique. To evaluate the robustness of LMs and highlight differences in their performance, we construct an adversarial test dataset that places the models under challenging conditions. Moreover, to address the lack of VP transcripts, we create transcripts by referencing existing or new types of VP techniques. We compare cases where evaluation criteria are included, the CoT technique is applied, or both are used together. In the experiment, our results show that the Llama3-8B model, fine-tuned with a dataset that includes a prompt with VP evaluation criteria, yields the best performance among small LMs and is comparable to that of a GPT-4-based VP detector. These findings indicate that incorporating human expert knowledge into the prompt is more effective than using the CoT technique for small LMs in VP detection.
\end{abstract}

\begin{CCSXML}
<ccs2012>
 <concept>
 <concept_id>10010147.10010178.10010179</concept_id>
 <concept_desc>Computing methodologies~Natural language processing</concept_desc>
 <concept_significance>500</concept_significance>
 </concept>
 <concept>
 <concept_id>10010405.10010455</concept_id>
 <concept_desc>Applied computing~Law, social and behavioral sciences</concept_desc>
 <concept_significance>300</concept_significance>
 </concept>
 <concept>
  <concept_id>00000000.00000000.00000000</concept_id>
  <concept_desc>Do Not Use This Code, Generate the Correct Terms for Your Paper</concept_desc>
  <concept_significance>100</concept_significance>
 </concept>
 <concept>
  <concept_id>00000000.00000000.00000000</concept_id>
  <concept_desc>Do Not Use This Code, Generate the Correct Terms for Your Paper</concept_desc>
  <concept_significance>100</concept_significance>
 </concept>
</ccs2012>
\end{CCSXML}

\ccsdesc[500]{Computing methodologies~Natural language processing}
\ccsdesc[300]{Applied computing~Law, social and behavioral sciences}
%\ccsdesc{Do Not Use This Code~Generate the Correct Terms for Your Paper}
%\ccsdesc[100]{Do Not Use This Code~Generate the Correct Terms for Your Paper}

\keywords{voice phishing detection, fine-tuning,  small language models}


\maketitle

\section{Introduction}
Voice phishing (VP), a type of fraud wherein scammers deceive victims over the phone to steal personal information and/or acquire financial gains, has become a pervasive crime worldwide. In the United States, financial losses from imposter scams amounted to \$2.7 billion in 2023, more than double the \$1.2 billion reported in 2020, with phone calls being the primary method of contact \cite{US2023}. 
In South Korea, the amount of financial damage caused by VP increased from \$31 million in 2018 to \$58 million in 2021 \cite{Seo2022}. The primary method for preventing VP involves educating individuals by announcing the characteristics and examples of such scams, thereby enabling them to recognize potential phishing attempts during phone calls. However, since scammers  try to impersonate investigative agencies or financial institution employees in a highly convincing manner, even working adults may find it difficult to recognize VP attempts.
In this context, governments worldwide are implementing various policies to reduce losses from VP.

One promising idea is to apply natural language processing (NLP) in identifying VP. The Korean government released hundreds of VP transcripts and call recordings to inform the public about the types and methods of VP scams. Leveraging this data, several VP detectors using NLP techniques have been proposed by research groups primarily based in Korea. These schemes used latent sementic analysis (LSA) and Doc2Vec \cite{kim2021voice}, a BERT model \cite{Milandu2022}, and combination of a BERT and various traditional machine learning classifiers \cite{yu2024korean}. To enhance the accuracy of VP detection, a proprietary generative large language model (LLM), such as OpenAI's GPT-3.5, was utilized \cite{Sim_2024}. 

Proprietary LLMs developed by major technology companies are capable of understanding and generating human-like, even professional-level, text. Moreover, thanks to emergent abilities of LLMs such as in-context learning, they do not require fine-tuning for specific tasks. Therefore, the introduction of proprietary LLMs to VP detectors leads to performance improvements compared with previous works as shown in our previous study \cite{Sim_2024}. 
However, since their architectures, training data, and parameter values are not publicly disclosed, they raise several concerns. First, the use of proprietary LLMs through 
application programming interfaces (APIs) incur high operational costs. Second, there is a risk of privacy breaches as private contents are transmitted to external servers. To address these limitations, both industry and academia are working to customize open-source, pre-trained, and smaller-scale language models (LMs) to achieve their specific objectives \cite{wang2024comprehensive}. While this approach may reduce the emergent abilities shown in LLMs and extend the development period, it can lower operational costs and address privacy concerns. We refer to this type of LM as an SLM (small language model).

In this paper, we fine-tune Llama3 (8B), a representative SLM, to serve as a VP detector. To enable an SLM to perform at a level comparable to proprietary LLMs, we incorporate well-designed VP evaluation criteria into the prompt and utilize the Chain-of-Thought (CoT) technique. In addition, we construct an adversarial dataset to test the
robustness of various VP detectors and highlight differences in their performance. Moreover, to address the lack of VP transcripts, we create transcripts by referencing existing and new types of VP techniques. We compare schemes where evaluation criteria are included, the CoT technique is applied, or both are used together. 
We also assess the impact of transcript block size. The main contributions of our paper are described in more detail as follows:

\begin{itemize}
\item By referring to an article analyzing VP techniques, we create 11 evaluation criteria for LMs to use when assessing the likelihood of VP. We demonstrate that including these evaluation criteria in the prompt significantly improves the performance of SLMs.

\item We create a set of transcripts that do not involve VP attempts but include vocabulary and sentences commonly found in phone scams. Using these transcripts along with those involing VP attempts, we construct an adversarial test dataset. This dataset can be used as a challenging benchmark for evaluating the performance of VP detectors.

\item We demonstrate that an SLM fine-tuned with training data incorporating prompts with VP criteria outperforms both the CoT-only approach and the approach that combines CoT and VP criteria. This suggests that, when core information crafted by human specialists is provided, the SLM can achieve performance comparable to proprietary LLMs, whereas CoT-approaches cannot.
\end{itemize}

% Our VP detection process is composed of two stage: first, VP possibilities of each block are obtained using the LLM, second,  making decision on the transcript using a weighted average of VP likelihood. 

The remainder of this paper is organized as follows. Section \ref{sec:related_work} reviews related work in the context of our research. In Section \ref{sec.collecting_data}, we describe the process of collecting and preprocessing the datasets. Section \ref{sec.proposed} presents the proposed VP detector. Numerical results are discussed in Section \ref{sec:results}. Finally, Section \ref{sec:conclusion} concludes the paper.  Collected dataset and source codes are available at: \url{https://github.com/kufany/VP_detector_SLM}.

\section{Related works}\label{sec:related_work}
\subsection{Exsiting VP detectors}
In the work by \citet{kim2021voice}, audio from phone calls was transcribed into text, and either LSA or Doc2Vec was applied to determine whether the conversation involved VP. The classification accuracies of both methods were below 0.61. \citet{Milandu2022} utilized the KoBERT model, a Korean adaptation of the BERT model, to detect VP. During data preprocessing, extraneous elements, such as special characters, punctuation, phone numbers, and Korean stop words, were removed, and tokenization was performed using a morphological analyzer. This KoBERT model achieved an accuracy of up to 99.6\%. \citet{yu2024korean} applied named entity recognition (NER) with key tags and sentence-level N-grams. They used an un-fine-tuned BERT model to convert groups of sentences (i.e., sentence-level N-grams) into vectors, which were then classified as either VP or non-VP using traditional machine learning techniques such as logistic regression (LR), decision tree, etc. Among these, the LR-based scheme achieved the highest accuracy of 97.3\%. In our previous work \cite{Sim_2024}, a variant of OpenAI's GPT-3.5, the text-davinci-003 model, was used for a VP detector. 
Due to the model's token limit and the variable length of transcripts, each transcript was divided into fixed-size blocks and each block was inserted into a prompt. Each prompt included 19 VP criteria to assist the LM in calculating the VP likelihood. The VP likelihoods of all blocks within a transcript were compared with two thresholds—a likelihood threshold and a frequency threshold—to determine whether the transcript indicates VP. Evaluated in five different scenarios using various sources of training and test data, the GPT-based detector yielded accuracies of 94–97\%, while the BERT-based detector yielded accuracies of 89–97\%.

%Other key differences from previous studies include the adoption of a fine-tuned SLM, an improved design of VP criteria, and the introduction of the CoT technique. 

%This work extends the work of \citet{Sim_2024}, in which GPT-3.5 was tasked with outputting the likelihood that a block of the transcript is related to voice phishing. The core detection process—converting call audio to a transcript, splitting it into multiple blocks, inputting each block into the LLM as part of a prompt, obtaining the LLM's output regarding the likelihood of voice phishing, and integrating these outputs to make a final decision—remains consistent with the methodology used in our prior study. 

\subsection{Base model - Llama 3}
%Open-source small language models (LLMs) are regarded as viable alternatives to proprietary LLMs, addressing issues related to high usage costs and potential privacy risks. While open-source LLMs are generally smaller in scale compared to proprietary counterparts, they can be fine-tuned to develop models tailored for specific tasks, achieving performance levels closer to those of proprietary LLMs. Presently, there is a growing trend across various industries to adopt open-source LLMs. 

Meta’s release of Llama models in various parameter sizes has driven significant progress in the open-source (OS) SLM ecosystem \cite{touvron2023llama}. In July 2024, Meta made publicly available a new set of foundation models, called Llama 3.1, supporting multilingual capabilities, reasoning, coding, and more. These models have been shown to perform comparably to leading proprietary models, such as GPT-4, across a wide range of tasks \cite{dubey2024llama3herdmodels}. Llama 3 models continue to be fine-tuned for use across multiple fields, such as the medical field \cite{adams2024llama} and mathematics \cite{zhang2024accessing}. 

Meanwhile, Korean-English bilingual language model based on Llama 3, named Bllossom model, was released enhancing the connection of knowledge between Korean and English \cite{bllossom}. With vocabulary expansion, instruction tuning for Korean culture, and vision-language alignment, the model delivers state-of-the-art performance among Korean OS SLMs and is widely used across various fields utilizing the Korean language. In this context, we use Llama-3-Korean-Bllossom-8B as our base model.

\subsection{Chain-of-Thought}
CoT refers to a sequence of progressive reasoning steps that enable LLMs to generate accurate solutions for reasoning tasks, such as arithmetic, commonsense, and symbolic reasoning \cite{wei2022chain}. A notable advantage of CoT lies in its implementation simplicity, requiring only a CoT-specific prompt without necessitating an increase in model size or dataset scale. However, these reasoning capabilities appear to emerge predominantly in models with at least tens of billions of parameters. 

Meanwhile, it has been demonstrated that smaller LMs (student models) can acquire task-specific reasoning capabilities from larger LMs (teacher models) through a CoT-based knowledge distillation \cite{magister2022teaching, ho2022large}. In this process, the student model is fine-tuned using datasets consisting of answers generated by teacher models in response to CoT-style prompts. 
In this paper, we try to adopt CoT-based knowledge distillation into SLM for the potential performance improvement of a VP detector.

%==============================================================
\begin{figure*}[h]
	\centerline{\includegraphics[width=13cm]{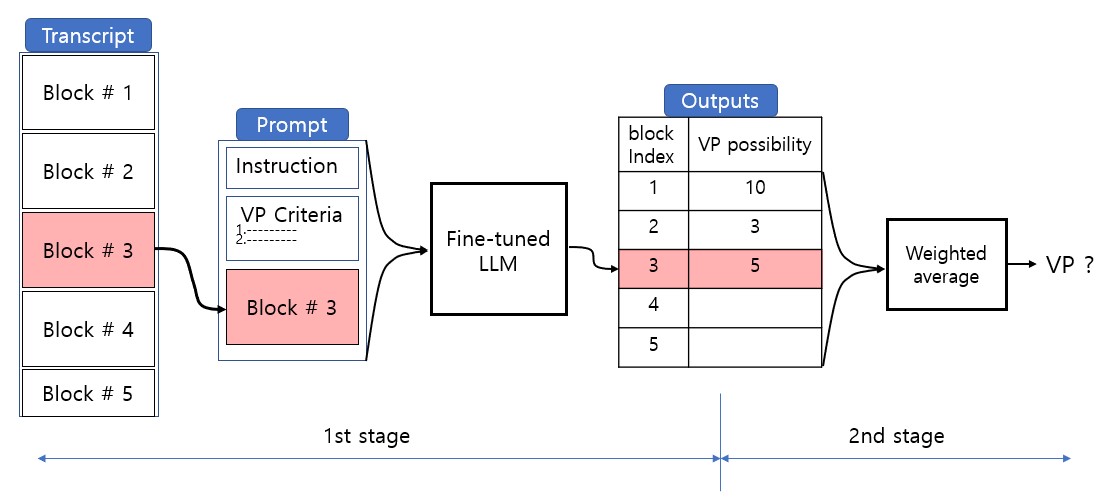}}
	\caption{The overall process of our VP detector.}
	\label{fg.system_model}
\end{figure*}
%==============================================================

\section{Data Collection}\label{sec.collecting_data}
\subsection{Transcripts obtained from data portals}

Table~\ref{dataset1} describes the topics, sources, and number of transcripts in various datasets used for training, validating and testing LMs. Dataset \textit{A} comprises 219 authentic VP transcripts, primarily obtained from \citet{Financial}. Some are converted from audio recordings, while others are originally in text format. Datasets \textit{B} to \textit{F} are non-VP datasets obtained from various data portals. Datasets \textit{B} and \textit{D}, which cover the areas of finance and insurance, are obtained from \citet{Hub} and contain 269 and 259 transcripts, respectively. Since financial-related terms are highly likely to appear in VP transcripts, utilizing these datasets for training may enhance the sensitivity of VP detectors. Datasets \textit{C}, \textit{E}, and \textit{F} consist of various everyday conversations obtained from \citet{corpus} and \citet{Hub}, containing 208, 147, and 170 transcripts, respectively. 

\begin{table}[t]
\centering
\caption{Topics of datasets}\small
\label{dataset1}
\begin{tabular}{lccc}
\toprule
 Class & Set & Topic  &  No. of transcripts \\
 \midrule
\multirow{2}{*}{VP} 
    & A   & Transcripts of actual VP calls \cite{Financial, KBS} &  219 \\  
    & H   & VP transcript created by human & 35\\
\midrule
\multirow{6}{*}{non-VP} 
    & B   & Finance counseling \cite{Hub} & 269 \\
    & C   & Everyday topics \cite{corpus} & 208\\
    & D   & Finance/insurance civil complaint \cite{Hub} & 259\\ 
    & E   & Everyday topics \cite{Hub}  & 165 \\
    & F   & Complaints regarding small businesses and the public in 10 sectors \cite{Hub} & 164\\
    & G   & Everyday conversation easily confused with VP (created) & 58\\
\midrule
\multicolumn{3}{c}{total} & 1377\\
\bottomrule
\end{tabular}
\end{table}

\subsection{Adversarial dataset}
As shown in previous studies \cite{Milandu2022, yu2024korean, Sim_2024}, existing methods have already achieved near-perfect accuracy. Most prior work used everyday conversations or civil complaint calls as non-VP samples. However, such conversations, which lack vocabulary or context resembling VP, are rarely mistaken for VP by the general public and therefore may not serve as suitable non-VP samples for practical evaluation. To address this issue, we propose the creation of transcripts that are highly likely to be misclassified as VP, thereby constructing an adversarial test dataset. The topics of these transcripts are as follows: 1) sharing personal VP experiences with acquaintances, 2) police summons notice due to involvement in a crime, 3) reporting VP to the police, 4) description of loan products, 5) verifying personal information for credit card delivery, etc. LMs with limited contextual understanding ability are more prone to misclassifying these transcripts as VP. This dataset is denoted by \textit{G} in Table \ref{dataset1}. Dataset \textit{G}, along with portions of datasets \textit{A} and \textit{H}, constitutes an adversarial test dataset. Selective samples of dataset \textit{G} are shown in Table \ref{tb.Dataset_G} of Appendx \ref{app.dataset_G}. 

\subsection{Approach to Address the Lack of VP Transcripts}
In VP detection, obtaining transcripts that contain real VP attempts is essential. However, such transcripts are rarely available to the public due to their sensitive nature, as they are typically part of criminal investigations and often contain personally identifiable information. Even in exceptional cases where transcripts are released, strict protocols and safeguards are required. To address this data scarcity, we generated synthetic VP transcripts. Initially, we explored using ChatGPT to produce VP conversations. However, the generated contents were unrealistic and coarse for practical use. Consequently, we manually created VP transcripts. These transcripts were designed based on common patterns observed in existing VP cases. Additionally, to reflect recent trends, we incorporated emerging VP tactics reported between 2020 and 2023 by sources such as \citet{NIS} and \citet{wise_user}. This process resulted in the creation of 35 virtual VP transcripts, and the resulting dataset is referred to as Dataset \textit{H}, as described in Table~\ref{dataset1}. 

Meanwhile, it is much easier to obtain non-VP transcripts, and we included a greater number of them in our dataset than VP transcripts. {In anomaly detection, class imbalance is a common issue, as abnormal data is typically much scarcer than normal data \cite{yang2024ad}.} In such settings, models are often trained and evaluated predominantly on normal samples. In our dataset, the ratio of VP to non-VP transcripts is approximately 1 to 4.4.

\subsection{Preprocessing}
During preprocessing, we refine the transcripts used for fine-tuning, aiming to enhance the LM's understanding of VP. The truncation of sentences and inaccurate expressions in a transcript stem from the screening of personal information in the original audio files and errors in the speech-to-text conversion process. To address these issues, human reviewers manually corrected awkward segments and filled in missing parts using reasonable estimates.

\section{VP detector}\label{sec.proposed}
The overall process of our VP detector, which receives a transcript and determines whether it constitutes an attempt at VP, is illustrated in Fig. \ref{fg.system_model}. The process is mainly divided into two stages. In the first stage, the transcript is divided into multiple blocks. This is because the length of the input-output sequence is limited in LMs, and overly long transcripts may not fit into them. Even though an LM allows long sequences, its ability to understand and generate may decrease if the sequence length exceeds certain limits. 
After that, the LM analyzes each block and outputs the VP likelihood, which is used as the label for each block. This method was originated in \cite{Sim_2024}, but we have improved the contents of the prompt, such as instructions and VP criteria, and applied CoT. In the 2nd stage, a final decision is made based on the weighted average of VP likelihoods across all the blocks. Each step of the process is described in detail in this section.

%The key differences from the process considered in work of \citet{Sim_2024} include improvements to the instructions and VP criteria, the use of a fine-tuned LLM, and the adoption of chain-of-thought (CoT) prompts and a weighted average VP likelihood. 

\subsection{Prompt design} \label{sec:prompt}
The prompt mainly consists of three parts: an instruction explaining what is being asked of the LM, the VP criteria, and a block of the transcript. An example prompt is shown in Table \ref{tb.prompt}. The first paragraph contains the instruction for the LM, where the underlined sentences represent the system's role, and the remaining sentences represent the user's role. This paragraph assigns the role of a VP detector to the LM, instructs it to refer to the VP evaluation criteria, and asks it to respond with an integer between 0 and 10. The reason for obtaining integers as output is to simplify the extraction of VP likelihood from the text generated by the LM. The content of the instruction changes when CoT is applied, as will be explained in a later section.

%%****************************************
\begin{table}[t]
\caption{An example of a prompt with VP criteria}
\label{tb.prompt}
    \centering
    \begin{tabular}{m{12cm}}
     \toprule
     \uline{You are a voice phishing detector. The evaluation criteria use logical operators such as "and" or "or."} 
     Referring to the evaluation criteria, review the transcript to determine whether it involves an attempt at voice phishing. Return the voice phishing likelihood as an integer between 0 and 10. Provide only the likelihood score and do not include any additional commentary. If it is determined to be an ordinary conversation, return `0'.
\newline 
\newline 
---Evaluation creteria---
\newline 
1.A financial institution employee offers a low-interest loan to the recipient. \\
$\quad\quad\quad\vdots$\\
11.The recipient is instructed to \{(hand over a debit card or account to ...
\newline 
-------------------------------\\
<Call transcript begins>
\newline
\newline
Participant 2: Hello?\\
Participant 1: Hello, this is Investigator Choi Man-ho from the Seoul Central District Prosecutors' Office. Is this Mr. Kim Chi-ho?\\
$\quad\quad\quad\vdots$
\newline
<Call transcript ends>
\\
     \bottomrule
    \end{tabular}
\end{table}
%%****************************************

In the second paragraph, the VP criteria are presented, with their details provided in Table~\ref{tb.VP_technique} of Appendix \ref{App.prompt}. In developing these criteria, we reviewed an article that analyzes VP types by crime stage and the characteristics associated with each stage \cite{StageofVP}. Based on this analysis, we identified 8 types of VP schemes, each of which progresses through 3 to 8 stages. However, including all crime stages for each type in the VP criteria would result in an excessively large prompt. To mitigate this, we identify common crime stages across different types, eliminate redundancies, and abstract the statements. Furthermore, we employ logical operators such as "and" and "or" to consolidate multiple crime stages into a single criterion. Consequently, the VP criteria are condensed into 11 sentences.

At the end, a block of a transcript is added. The length of each block, i.e., the number of letters in each block, is limited to a specific value, with all blocks having identical lengths except for the last one. 
The length of the last block may be shorter than the limit. Sentences can be split across different blocks, which could result in a loss of meaning. However, due to the LM's ability to complete incomplete sentences, we believe the loss would be marginal. 
%==============================================================
\begin{figure}[h]
	\centerline{\includegraphics[width=11cm]{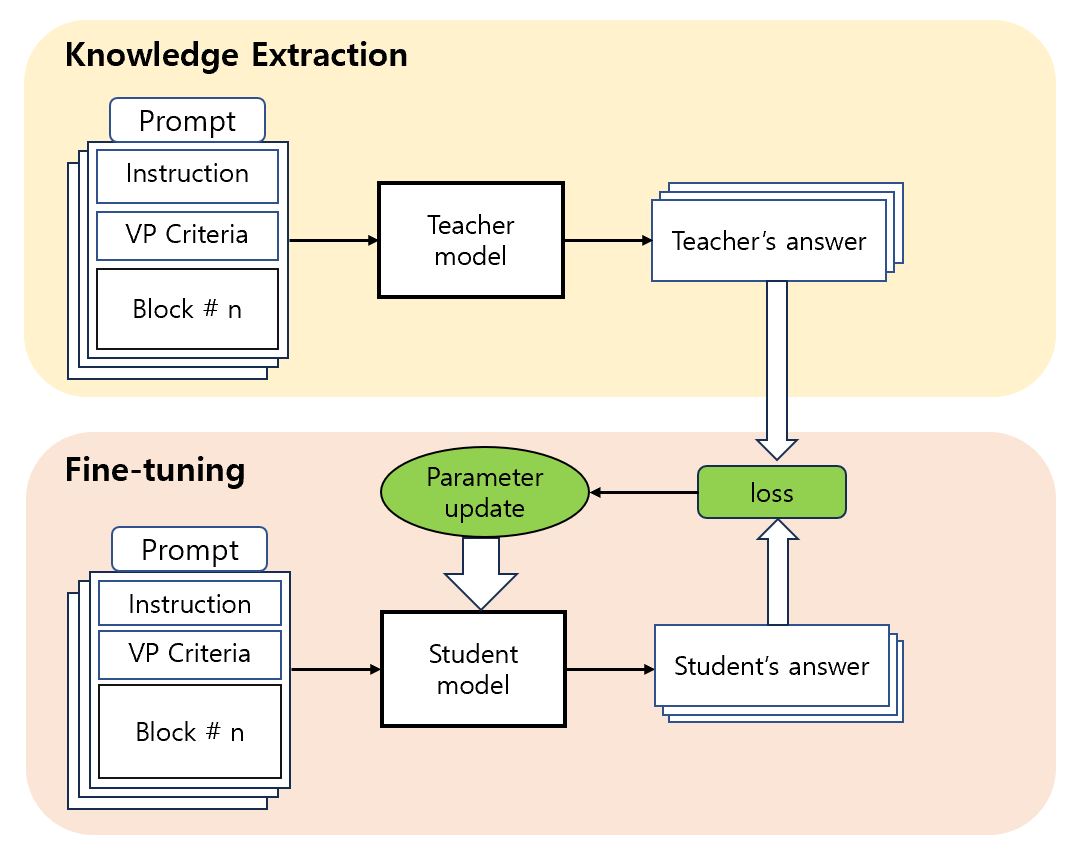}}
	\caption{Fine-tuning process.}
	\label{fg.fine_tuning}
\end{figure}
%==============================================================

\subsection{Fine-tuning}
Fig. \ref{fg.fine_tuning} illustrates the process of supervised fine-tuning of the SLM. Since manual labeling requires significant time and effort, we assign the labeling task to GPT-4o, which can be regarded as a form of knowledge extraction where the teacher model is GPT4o and teacher's answer is VP likelihood. The prompt explained in the previous subsection is used to obtain the answer from GPT4o. This knowledge extraction is completed before fine-tuning. The student model, the Llama-3-Korean-Bllossom-8B in our case, is fine-tuned with the prompt and the labels obtained in the knowledge extraction. Table \ref{tb.no_files} describes the number of VP or non-VP transcripts used as training, validation, test datasets for the 1st and 2nd stages, respectively. Since the lengths of VP transcripts are generally longer than those of non-VP transcripts, the ratio between the number of blocks corresponding to VP and that corresponding to non-VP becomes approximately 1:2 in the first stage.

We apply parameter-efficient fine-tuning using low-rank adaptation (LoRA) with a rank of 8 and 8-bit quantization~\cite{hu2021lora}, with fine-tuning conducted over 5 epochs. The length of the blocks included in prompts for fine-tuning is fixed at 500 to reduce API usage costs, while we test the LM with various block-length values. 

\begin{table}[h]
    \centering
    \begin{tabular}{m{0.7cm}|c||c|c|c|c}
        \toprule
        \multicolumn{2}{r||}{stage} & \multicolumn{2}{c|}{1st} & \multicolumn{2}{c}{2nd} \\ \cline{3-6}
        \multicolumn{2}{l||}{topic}  & Train & Val & Val & Test \\ \hline
        \multicolumn{2}{c||}{VP}  & 159  & 41 & 41 & 54 \\ \hline
        \multirow{2}{=}{\centering Non-VP} & Normal & 681 & 168 & 41 & 216 \\ \cline{2-6}
        & Adversarial & - & - & - & 58 \\ 
        \bottomrule
    \end{tabular}
    \caption{The number of transcripts used in different stages.}
    \label{tb.no_files}
\end{table}

%\subsection{Training and validation}\label{sec:train_val}

\subsection{Inclusion of CoT}
We apply CoT-based fine-tuning to our model. Several recent studies have reported that CoT-based fine-tuning improves the performance of SLMs on reasoning tasks such as arithmetic reasoning, commonsense reasoning, and symbolic reasoning \cite{magister2022teaching, ho2022large}. However, the benchmarks commonly used in those studies are not as complex as in VP detection. We note that the introduction of CoT may not necessarily lead to improved results in SLM-based VP detection.

We consider two types of CoT prompts: (1) a prompt that instructs the generation of a CoT answer without including the VP criteria, and (2) a prompt that incorporates the VP criteria and instructs the generation of a CoT answer. The prompts of the first and second types are referred to as the CoT-type and CoTCri-type prompts, respectively, and described in Table \ref{tb.CoT_prompt} of the Appendix.

\subsection{Second stage} \label{sec:second_stage}
In the second stage, The outputs of the LM are weighted and averaged based on the length of the blocks. The weighted average of VP likelihoods, $\bar{P}$, is defined as 
\begin{align}
\bar{P} = \sum\limits_{i=1}^{N_{\text{block}}} \frac{l_i}{L} \times P_i,
\end{align}
where $l_i$ and $P_i$ denote the number of letters and the VP likelihood of the $i$-th block, respectively, and $N_{\text{block}}$ and $L$ denote the total number of blocks and the total number of letters in a transcript, respectively. The average VP likelihood is compared with a threshold $\lambda$ to determine whether the transcript is VP or not. We select $\lambda$ such that maximizing the accuracy evaluated on the validation dataset of the 2nd stage. The VP transcripts in the validation sets of the second stage are the same as those in the first stage, and the non-VP transcripts in the validation sets of the second stage come from those in the first stage.

\section{Experimental Results}\label{sec:results}
In this section, we evaluate the accuracy performance of the GPT-4, Llama-base, fine-tuned Llama, and KoBERT models on both the normal test dataset and the adversarial test dataset. The Llama-base model, which is not fine-tuned, is used for the ablation study. KoBERT denotes the scheme proposed by \citet{Milandu2022}. We use the following convention to label the different LMs and fine-tuning methods: \textit{X-Y-Z}, where \textit{X} represents the name of the LM, \textit{Y} specifies whether fine-tuning is applied (Base: not applied, FT: fine-tuning applied), and \textit{Z} indicates the content included in the prompt (Cri: VP criteria only, CoT: CoT only, CoTCri: both VP criteria and CoT, Plain: neither VP criteria nor CoT).

\begin{table*}[h]
\caption{Accuracy of Voice Phishing Detection Using GPT-4, Llama3, and KoBERT on a normal test dataset.}
    \label{tb:accuracy1}
    \centering
    \begin{tabular}{l||c|c|c|c|c|c}
    \toprule
    \textbf{Block-length} & \textbf{100} & \textbf{300} & \textbf{500} & \textbf{900} & \textbf{1500} & \textbf{2500} \\ \midrule    
    GPT4o-CoT & - & 96.30 & 98.15 & 99.26 & 97.78 & - \\
GPT4o-Cri & - & 97.78 & 99.26 & 98.89 & \hl{100} & - \\
GPT4o-CoTCri & - & 98.89 & 98.15 & 98.15 & 97.41 & - \\ \midrule
LLAMA3-Base-Plain &  83.33 & 82.59 & 69.63 & 74.07 & 59.63 & 40.0 \\
LLAMA3-Base-CoT & 79.63 & 75.56 & 64.07 & 65.93 & 59.63 & 61.11 \\
LLAMA3-Base-Cri & 88.15 & 88.89 & \hl{91.11} & 84.44 & 84.81 & 75.56 \\
LLAMA3-Base-CoTCri & 80.74 & 83.33 & 89.63 & \hl{91.11} & 86.30 & 87.41 \\ \midrule
LLAMA3-FT\_Plain & 98.15 & 98.89 & 99.26 & \hl{100} & 99.63 & 99.63 \\
LLAMA3-FT-CoT & 99.26 & 98.52 & 99.26 & 99.63 & \hl{100} & \hl{100} \\
LLAMA3-FT-Cri & 98.89 & 99.63 & \hl{100}& \hl{100} & \hl{100} & 99.63 \\
LLAMA3-FT-CoTCri & 93.33 & 98.15 & 98.52 & 98.15 & 96.67 & 97.04 \\ \midrule
      KoBERT & \multicolumn{6}{c}{98.52} \\
\bottomrule
    \end{tabular}    
\end{table*}

\subsection{Normal Test Dataset}
Table \ref{tb:accuracy1} summarizes the accuracies 
across different models, fine-tuning methods, and prompt contents, tested on the normal dataset by varying the block-length from 100 to 2500. We first focus on the Llama3-Base schemes. It is observed that Llama3-Base-Cri and Llama3-Base-CoTCri achieve the highest accuracy of 91.11\% when the block-length is set to 500 and 900, respectively. On the other hand, Llama-Base-CoT exhibits the lowest accuracy overall. This indicates that incorporating CoT solely in the prompt may degrade performance in non-fine-tuned SLMs. Additionally, LLaMA3-Base-Plain achieves its peak accuracy of 83.33\% at a block length of 100, exhibiting inferior performance compared to LLaMA3-Base-Cri. This suggests that including VP criteria in the prompt provides a performance improvement. Lastly, it is observed that as the block-length increases, the overall performance tends to decrease.

Examining the results of the Llama3-FT schemes, it is evident that its overall performance is significantly improved compared to Llama3-Base schemes, achieving high accuracies exceeding 97\% in all cases except for Llama-FT-CoTCri. LLaMA3-FT-Cri achieves perfect accuracy (100\%) at block lengths of 500, 900, and 1500. Llama-FT-CoT achieves an accuracy of 100\% when the block-length is 1500 and 2500. However, LLaMA3-FT-CoTCri consistently yields the lowest accuracy across all block lengths, with a peak value of 98.52\% at a block length of 500. KoBERT achieves an accuracy of 98.52\%, while the GPT4o schemes achieves the highest accuracy of 100\%. \cite{Milandu2022}

\begin{figure}[h]
	\centerline{\includegraphics[width=11cm]{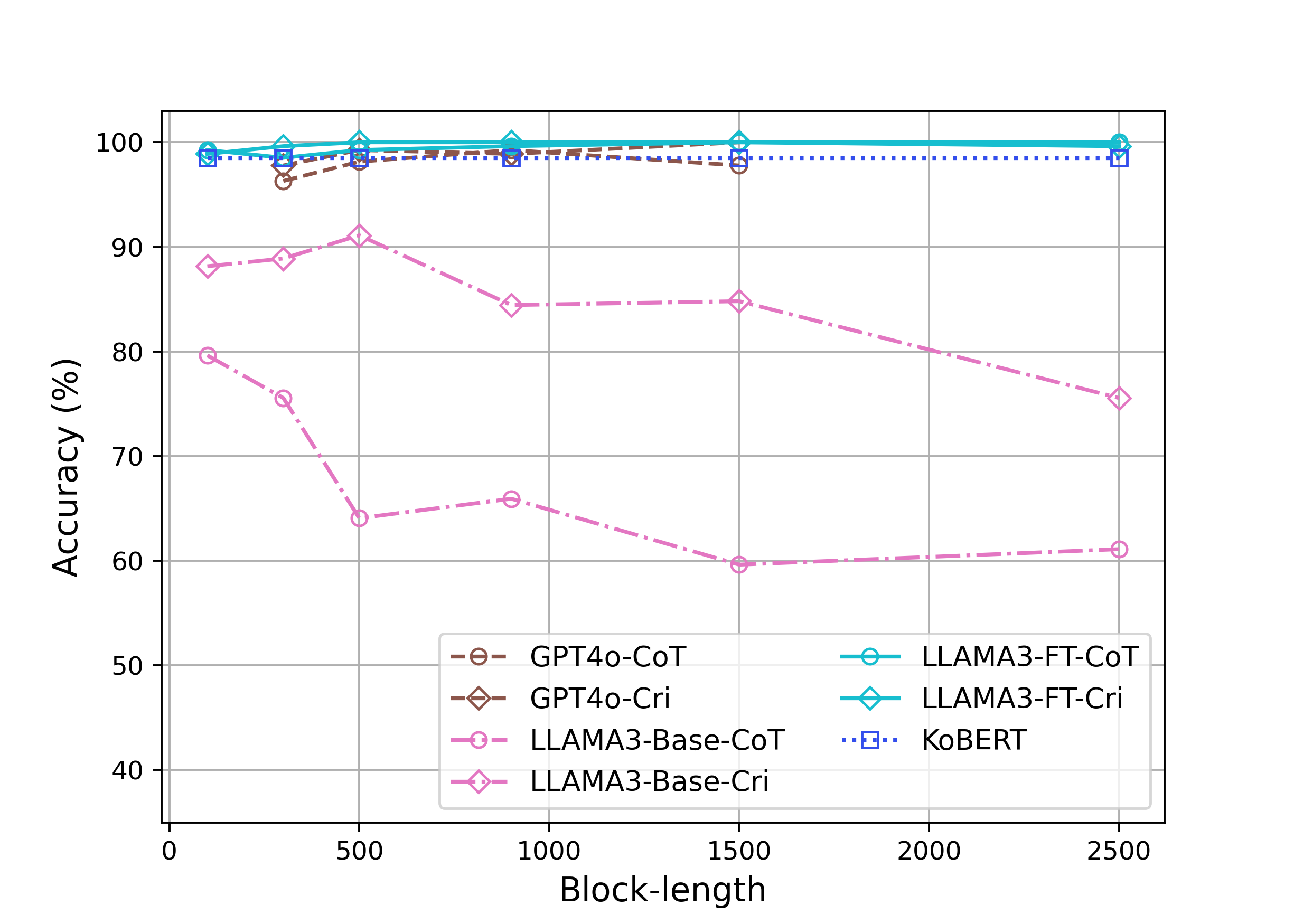}}
	\caption{Accuracy across varying block-length for GPT4o, Llama3-Base, Llama3-FT, KoBERT schemes except for Plain and CoTCri type prompts on the normal dataset.}
	\label{fg.result1}
\end{figure}

Fig. \ref{fg.result1} shows the accuracies accross varying block-lengths for the GPT4o, Llama3-Base, Llama3-FT, KoBERT schemes except for Plain and CoTCri type prompts. We observe that the GPT4o, Llama3-FT, KoBERT schemes achieve accuracies close to 100\%, while the Llama3-Base scheme exhibits inferior performance as the block length increases.
These results provide insufficient evidence to establish a clear superiority among the techniques. This emphasizes the need for the challenging test dataset. 

%==============================================================
%==============================================================

\subsection{Adversarial Test Dataset}
\begin{table*}[h]
    \caption{Accuracy of Voice Phishing Detection Using GPT-4, Llama3, and KoBERT on an adversarial test dataset.}
    \label{tb:accuracy2}
    \centering
    \begin{tabular}{l||c|c|c|c|c|c}
    \toprule
    \textbf{block-length} & \textbf{100} & \textbf{300} & \textbf{500} & \textbf{900} & \textbf{1500} & \textbf{2500} \\ \midrule
    GPT4o-CoT & - & 85.71 & 87.50 & 91.96 & 91.07 & - \\
GPT4o-Cri & - & 90.18 & 95.54 & 96.43 & \hl{99.11} & - \\
GPT4o-CoTCri & - & 76.79 & 76.79 & 85.71 & 89.29 & - \\\midrule
    LLAMA3-Base-Plain & 78.57 & 83.04 & 78.57 & 71.43 & 57.14 & 54.46\\
LLAMA3-Base-CoT & 80.36 & 75.00 & 70.54 & 71.43 & 64.29 & 63.39 \\
LLAMA3-Base-Cri & 79.46 & 86.61 & \hl{88.39} & 76.79 & 75.00 & 70.54 \\
LLAMA3-Base-CoTCri & 64.29 & 71.43 & 66.96 & 76.79 & 67.86 & 66.96 \\ \midrule
   LLAMA3-FT-Plain & 83.04 & 82.14 & 91.07 & 83.93 & 84.82 & 87.50 \\
LLAMA3-FT-CoT & 75.89 & 75.89 & 81.25 & 81.25 & 83.93 & 83.04 \\
LLAMA3-FT-Cri & 91.07 & 91.07 & \hl{94.64} & 87.50 & 89.29 & 88.39 \\
LLAMA3-FT-CoTCri & 69.64 & 80.36 & 78.57 & 86.61 & 86.61 & 84.82 \\ \midrule
    KoBERT & \multicolumn{6}{c}{56.25} \\
    \bottomrule
    \end{tabular}
\end{table*}

Table \ref{tb:accuracy2} summarizes the accuracies over various schemes tested on the adversarial test dataset. In all schemes, it is obvious that the accuracies on the adversarial dataset are generally lower compared to the results on the normal dataset. The GPT4o-Cri scheme, with the block-length of 1500, achieves the highest accuracy of 99.11\% among all schemes. We can observe that the accuracies of GPT4o schemes tends to increase as the block-length increase. This can be attributed to the fact that the larger blocks include more information, so the high-capacity LLMs such as GPT4 are able to analyze the block more accurately. 

Among the Llama3-Base schemes, Llama3-Base-Cri achieves the highest accuracy of 88.39\% when the block-length is 500. This represents a gain of 5.35 \%p (Llama3-Base-Cri - Llama3-Base-Plain) as a result of incorporating VP criteria. In contrast, Llama3-Base-CoT and Llama3-Base-CoTCri show no significant improvements over Llama3-Base-Plain, and their performance is sometimes observed to degrade further. We can observe that the accuracies of Llama3-Base schemes tends to decrease as the block-length increase. This is in contrast to GPT-4, and it is presumed that SLMs like Llama 3 have limited capacity, resulting in reduced ability to analyze the larger blocks of text.

In the case of the Llama3-FT schemes, Llama3-FT-Cri achieves the highest accuracy of 94.64\% at a block-length of 500, which represents a gain of 6.25 \%p compared to Llama3-Base-Cri while an accuracy that is 4.47\%p lower than GPT-4-Cri. A notable characteristic of Llama3-FT-CoT and Llama3-FT-CoTCri is the tendency for accuracy to increase as the block-length grows. This suggests that the fine-tuning of CoT aids SLMs in analyzing larger amounts of information. Nevertheless, Llama3-FT-CoT and Llama3-FT-CoTCri exhibit lower performance than Llama3-FT-Cri, indicating that providing the VP criteria is much more beneficial for SLMs than applying CoT. KoBERT achieves an accuracy of 56.25\%, which represents a decrease of approximately 42 \%p compared to its performance on the normal dataset.

\begin{figure}[h]
	\centerline{\includegraphics[width=11cm]{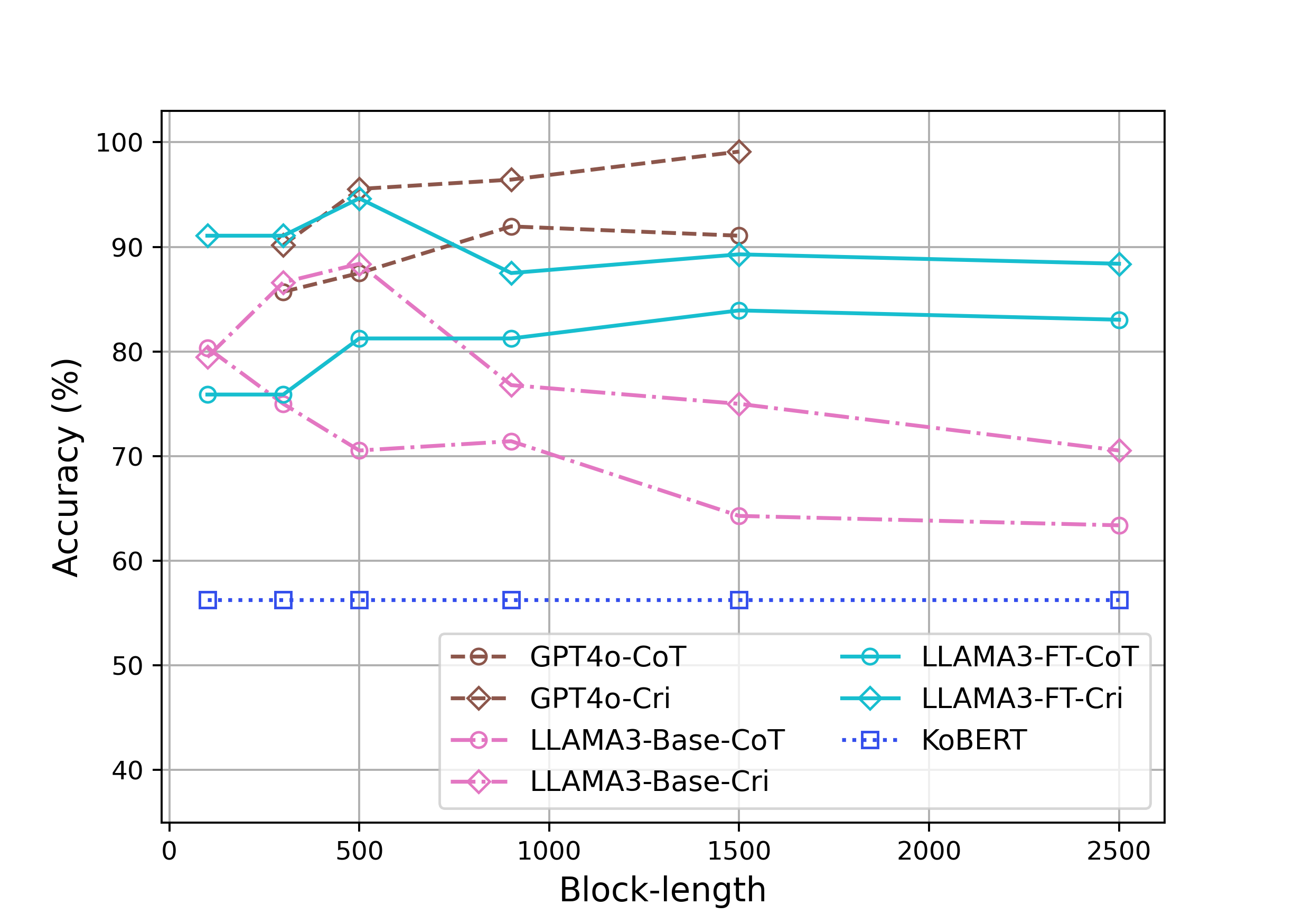}}
	\caption{Accuracy across varying block-length for GPT4o, Llama3-Base, Llama3-FT, KoBERT schemes except for Plain and CoTCri type prompts on the adversarial dataset}
	\label{fg.result2}
\end{figure}

Fig. \ref{fg.result2} describes the accuracies at different block-lenghts for the GPT4o, Llama3-Base, Llama3-FT, KoBERT schemes except for Plain and CoTCri type prompts on the adversarial dataset. It can be observed that the accuracy of the Llama3-FT-Cri scheme approaches that of GPT4o-Cri, while significantly outperforming the Llama3-FT-CoT, Llama3-Base, KoBERT schemes. 
%==============================================================

\section{Conclusion}\label{sec:conclusion}
We developed a VP detector by fine-tuning  a representative SLM, Llama3-8B. The proposed VP detector incorporated well-designed VP criteria in the prompt and/or adopted CoT. To enable rigorous evaluation across various LMs, an adversarial test dataset was constructed. Additionally, to address the scarcity of VP transcripts, we created transcripts by referencing both existing and novel VP techniques. The accuracy performance of GPT-4, Llama3-Base, Llama3-FT, and KoBERT models was evaluated using different prompt designs on both a normal dataset and an adversarial dataset. On the normal dataset, the GPT-4o, Llama3-FT, and KoBERT schemes achieved accuracies higher than 98\%. On the adversarial dataset, the Llama3-FT-Cri scheme demonstrated performance close to that of the best GPT4o-based scheme at a block length of 500, outperforming other schemes. Our findings also suggest that incorporating human expert knowledge into the prompt is more effective than using the CoT technique for small LMs in VP detection. Research on fine-tuning SLMs with significantly fewer parameters than the Llama3B model could provide insights into implementing a VP detector capable of running on mobile devices, making it an intriguing and challenging topic for future research.

%\begin{acks}
%To Robert, for the bagels and explaining CMYK and color spaces.
%\end{acks}

\bibliographystyle{ACM-Reference-Format}
\bibliography{sample-manuscript_arxiv}

\appendix
\section{Samples of Adversarial Test Dataset}\label{app.dataset_G}
Table \ref{tb.Dataset_G} shows part of the transcripts from Dataset G, with only key parts included due to space limitation. Names and places are fictional, and the original Korean transcripts were translated into English.

\begin{table*}[]
\caption{Samples of Dataset \textit{G}}
\label{tb.Dataset_G}
\centering
\begin{tabular}{m{1cm} | m{13cm} }
\toprule
Type & Content of transcripts \\
 \midrule
    sharing personal VP experiences & Participant 1: Taehyun, can you talk?

Participant 2: Sure, what’s up?

Participant 1: I had a scary experience today. A stranger called and offered me money if I let them use my bank account.

Participant 2: That sounds like voice phishing.

Participant 1: At first, I was tempted. I’ve been stressed about money lately. But their tone was odd, and they kept pushing me.

Participant 2: That’s a big red flag. They target people in tough situations. What did you do?

Participant 1: I asked questions to buy time, like “How would you send the money?” Then they told me to deposit money into a weird account. That’s when I knew—it was phishing. So I hung up.

Participant 2: Good move. I had a similar call once and reported it to the police. Now I don’t answer unknown numbers.
\\
         \midrule
     police summons notice due to involvement in a crime  & 
Participant 1: Hi, is this Jinwoo Kim?

Participant 2: Yes, who’s this?

Participant 1: This is Officer Minho Lee from the Dongnae Police Department in Busan.
We got a report today that your bank account might be connected to a fraud case.
We need you to come in for questioning.

Participant 2: What? My account? That can’t be right. There must be some mistake.

Participant 1: We just need to look into it.
We’ll go over all the details during the investigation.

Participant 2: Okay, I’ll come.
But I really don’t get how my account got involved in something like this.

Participant 1: From what we know, your account might’ve been used to move illegal funds.
We’ll figure out what really happened, but we need your help.

Participant 2: That’s so strange. I’ve never done anything like that…
When should I come in?

Participant 1: As soon as possible would be best.
Can you come by tomorrow at 10 a.m.?

Participant 2: Sure, I can do that.
Do I need to bring anything?

Participant 1: Just your ID and a copy of your recent bank transactions.

Participant 2: Got it. I’ll bring everything.
This is really surprising and a bit scary...

Participant 1: Don’t worry. As long as you tell the truth, we can clear things up.
Thanks for cooperating.
\\
     \midrule

reporting VP to the police,

& Participant 1: Hello, this is the Cyber Crime Division. How can I help you?

Participant 2: Hi, is this the police cyber unit?

Participant 1: Yes, that’s right. What can I help you with?

Participant 2: I think I got a voice phishing call recently, and I wanted to check if it really was.

Participant 1: I see. What kind of call did you get?

Participant 2: At first, they said it was a low-interest loan offer.
But during the call, they gave me a link to a website and told me to go there.
I clicked it, downloaded something, and ran it.
Then it showed a screen asking for my bank security card numbers.
That’s when I got suspicious. I’d also like to report it.

Participant 1: From what you’re saying, that definitely sounds like a voice phishing attempt.
Did you enter any personal information or card numbers?

Participant 2: No, I didn’t enter anything. But I did download and install the app they told me to.
It felt kind of shady.

Participant 1: In that case, you should delete the app right away.
Also, contact your phone carrier or visit their customer center for help.
You might even consider resetting your phone to be safe.

Participant 2: Got it. I’ll do that. Thanks for the advice.

Participant 1: No problem. Just be careful with unknown calls or messages.
And if someone tells you to visit a certain link or website, that’s often a red flag for phishing.
Never follow those instructions.

Participant 2: I understand. 
\\ \midrule

    \bottomrule
\end{tabular}
\end{table*}

\section{Prompts}\label{App.prompt}

Table \ref{tb.CoT_prompt} describes prompts for CoT and CoTCri approaches. The yellow-highlighted sentences are intended for generating a CoT answer, while the green-highlighted sentences are designed to instruct the model to refer to the VP 

\begin{table*}[]
\caption{Voice Phishing Evaluation Criteria}
\label{tb.VP_technique}
\centering
\begin{tabular}{m{0.3cm} | m{14cm} }
\toprule
No. & The content of each criterion \\
 \midrule
     1 & A financial institution employee offers a low-interest loan to the recipient.\\
         \midrule
     2  & A law enforcement official informs the recipient that their account has been used for criminal activities and requires investigation.\\
     \midrule
     3  & An employee of a stock loss compensation company proposes compensation for investment losses.\\
     \midrule
     4  & During a call, the employee asks the recipient to provide information about \{(existing loan details) or (account balance) or (bank name associated with previous transactions)\}.\\
     \midrule
     5  & The employee requests \{(an investigation over the phone regarding criminal charges) or (inputting information into an institutional website) or (installing an app)\}\\
     \midrule
     6  & The employee conditions \{(approval for a low-interest loan) or (an increase in loan limits)\} on \{(repayment of the recipient's existing loan) or (taking a new high-interest loan and repaying it immediately) or (notarizing income for a third party)\}.\\
     \midrule
     7  & For the purpose of \{(preventing further damage) or (conducting an investigation)\}, the recipient is asked to \{(withdraw savings) or (take out a new loan)\}.\\
     \midrule
     8 & The recipient is instructed to \{(meet another employee to hand over cash directly) or (store it in a locker) or (transfer it to a specific account)\}.\\
     \midrule
     9 & The employee directs the recipient to \{(make payments with gift cards) and (share the gift card identification numbers)\} for investigative purposes.\\
     \midrule
     10 & The recipient is asked to transfer money to a specific account with the promise of a highly profitable investment.\\
     \midrule
     11 & The recipient is instructed to \{(hand over a debit card or account to another person) or (open a bank account under another person's name)\}.\\
    \bottomrule
\end{tabular}
\end{table*}

%%%***********************************************
\begin{table*}[]
\caption{Prompts of CoT and CoTCri approaches.}
\label{tb.CoT_prompt}
    \centering
\begin{tabular}{m{0.6cm}| m{14cm}}
\toprule
types & Prompt \\
\midrule
CoT & You are a voice phishing detector. \hl{Respond using the Chain-of-Thoughts method.} Review the call transcript and determine whether it is a voice phishing attempt. Provide a voice phishing likelihood score as an integer between 0 and 10. \hl{Briefly explain the reasoning behind your judgment, including a concise summary of relevant parts of the transcript that support your conclusion, within 200 characters.} End your response with: "Therefore, the likelihood is [ ]." If there is no sign of phishing, write "Therefore, the likelihood is [0]."
\newline
\newline
<Call transcript begins \& ends>
\newline
\hl{Let's think step-by-step,}  \\\midrule
CoT-Cri
& 
You are a voice phishing detector. \hl{Respond using the Chain-of-Thoughts method}. In the evaluation criteria, and and or represent logical operators. \sethlcolor{green}\hl{Refer to the evaluation criteria to assess the call transcript and determine if it is a voice phishing attempt.}\sethlcolor{yellow}  Provide a voice phishing likelihood score as an integer between 0 and 10. \hl{Briefly explain the reasoning behind your judgment, including a concise summary of relevant parts of the transcript that support your conclusion within 200 characters.} End your response with: "Therefore, the likelihood is [ ]." If there is no sign of phishing, write "Therefore, the likelihood is [0]."
\newline
\newline
-- Evaluation Criteria --
\newline
$\vdots$
\newline
<Call transcript begins \& ends>
\newline
\hl{Let's think step-by-step,}
\\
\bottomrule
\end{tabular}
\end{table*}
%%%**************************************

\end{document}